\documentclass[final,5p,times,twocolumn]{elsarticle}

\usepackage{lineno}
\usepackage{comment}
\usepackage{amsmath}
\usepackage{float}
\usepackage{graphicx}
\usepackage{svg}
\usepackage{pdfpages} 
\modulolinenumbers[5]
\usepackage{pgfplots}
\usepackage{hyperref}
\usepackage{amssymb}
\usepackage{multirow}
\usepackage{ gensymb }
\usepackage{multicol}
\usepackage{makecell}
\usepackage{caption}
\journal{arXiv}

\begin{document}
\begin{frontmatter}

\title{ \huge Deep brain state classification of MEG data}

\author{Ismail Alaoui Abdellaoui}
\author{Jesus Garcia Fernandez} 
\author{Caner Sahinli}
\author{Siamak Mehrkanoon\corref{cor1}}

\cortext[cor1]{Corresponding author}

\address{Department of Data Science and Knowledge Engineering, Maastricht University, The Netherlands}

\begin{abstract}

Neuroimaging techniques have shown to be useful when studying the brain’s activity. This paper uses Magnetoencephalography (MEG) data, provided by the Human Connectome Project (HCP), in combination with various deep artificial neural network models to perform brain decoding. More specifically, here we investigate to which extent can we infer the task performed by a subject based on its MEG data. Three models based on compact convolution, combined convolutional and long short-term architecture as well as a model based on multi-view learning that aims at fusing the outputs of the two stream networks are proposed and examined. These models exploit the spatio-temporal MEG data for learning new representations that are used to decode the relevant tasks across subjects. In order to realize the most relevant features of the input signals, two attention mechanisms, i.e. self and global attention, are incorporated in all the models. The experimental results of cross subject multi-class classification on the studied MEG dataset show that the inclusion of attention improves the generalization of the models across subjects.
\end{abstract}

\end{frontmatter}

\section{Introduction}

Functional neuroimaging is a collection of methods employed to extract information about the functioning of the brain by registering electromagnetic activity. Magnetoencephalography (MEG) is a noninvasive technique for investigating neuronal activity in the living human brain. In MEG studies, the weak 10 femtotesla — 1 picotesla magnetic fields produced by electric currents flowing in neurons are measured with multichannel SQUID (Superconducting Quantum Interference Device) gradiometers. MEG scans allow neuroscientists to study interesting properties of the working human brain, including spontaneous activity and signal processing following external stimuli \cite{hamalainen1993magnetoencephalography}.

Recent interest in machine learning techniques have sparked a surge in the development of data driven computational models to automatically learn the underlying nonlinear patterns of the data. The data driven based models have been successfully applied in many research fields ranging from dynamical systems, climate modelling, healthcare, biomedical signal analysis among others \cite{mehrkanoon2012approximate,mehrkanoon2015learning,mehrkanoon2014parameter,mehrkanoon2019deep,mehrkanoon2019deep2,mehrkanoon2018deep,davatzikos2019machine,breiman2001random,webb2018deep}. In particular, for biomedical signal and functional neuroimaging data, as stated in \cite{davatzikos2019machine}, earlier studies were focused on support vector machines \cite{davatzikos2019machine,lao2004morphological}, and later on random forests methods \cite{breiman2001random}. Machine learning techniques have been successfully employed to reveal neurological patterns for a number of diseases and disorders, such as Alzheimer's Disease \cite{kloppel2008automatic,zhang2011multimodal,rathore2017review}, brain development and aging \cite{franke2010estimating,habes2016advanced,xia2018linked}, preclinical states \cite{davatzikos2009longitudinal}, schizophrenia \cite{davatzikos2005whole}, mood disorders \cite{koutsouleris2015individualized} and autism \cite{ecker2010investigating}. While there has been extensive research into human activity classification using EEG (electroencephalography) data \cite{lawhern2018eegnet,sabour2017dynamic,zhang2018cascade,schirrmeister1703deep,roy2019deep,garg2017using}, there are less published results in the literature for MEG data.

Classifying human activity from functional brain data has been mainly studied as an application for developing BCI (brain-computer interface) software. In particular, deep learning models have shown to be effective in predicting human intentions and actions from EEG data. Within the state of the art, we can point out, e.g. the work in \cite{ha2019motor} where a capsule network\cite{sabour2017dynamic} (CapsNet) is leveraged to learn various properties of EEG signals. The authors in \cite{lawhern2018eegnet}, introduced a compact convolutional neural network for EEG-based BCIs. Finally, \cite{zhang2018cascade} aims to capture separately the temporal and spatial characteristics of EEG waves, for which an RNN architecture is combined with a CNN model in both a cascade and a multiview setting. 

This paper is organized as follows. A brief overview of the existing machine learning methodologies used to perform decoding from brain signals is given in Section \ref{sec:related_work}. A formal description of the two attention mechanisms used is presented in Section \ref{sec:preliminaries}. The proposed models are  introduced in Section \ref{sec:proposed_models}. The studied dataset is introduced in Section \ref{sec:data_desc}. The experimental results are reported in section \ref{sec:results}. Finally, a discussion followed by the conclusion are drawn in sections \ref{sec:discussion} and \ref{sec:conclusion}, respectively.
\section{Related Work}\label{sec:related_work}

Decoding human brain signals to infer the intention of the subject is an active inter- and multi-disciplinary field of research. Several methods have been introduced in the literature to interpret signals captured by neuroimaging techniques such as EEG, fMRI and MEG. For instance, the authors in \cite{waldert2008hand} predicted the direction of hand-movement from 9 right-handed subjects with the Regularized Linear Discriminant Analysis (RLDA) method using a combination of EEG and MEG signals. In another study by Friston et al. \cite{henson2011parametric}, they researched how to integrate multiple modalities, restrictions and subjects as to achieve higher reproducibility of cortical response across subjects. In their study, 18 subjects were shown faces from three different sets (famous, unfamiliar and scrambled faces) in three different sessions, each capturing the brain-signals either through MEG, EEG or fMRI. This inspired the following study \cite{olivetti2014meg} where machine learning and Transductive Transfer Learning (TTL) has been used to train and predict the faces seen by the subjects.

Another paper that was an inspiration to this study is the presentation of two deep learning architectures in order to enhance intention recognition based on EEG signals \cite{zhang2018cascade}. In their analysis, high precision in cross-subject classification tests is achieved with fusion of recurrent and convolutional components \cite{zhang2018cascade}. The work carried out in \cite{lawhern2018eegnet} proposes an EGG-based neural architecture that can extract interpretable features from neurophysiological phenomena. As a result, it has been clearly shown that the network performance is not driven by noise or artifact signals in the data \cite{lawhern2018eegnet}.
\section{Preliminaries}\label{sec:preliminaries}
The introduced models in this paper are equipped with two types of attention mechanisms: self and global. Attention mechanism, a technique that is introduced to capture long range interactions, is applied to the networks presented in section \ref{sec:proposed_models}, allowing them to learn to focus on specific parts of the brain's signals \cite{bello2019attention}. As it is shown in \cite{bello2019attention}, self-attention can potentially improve overall performance of the various networks \cite{bello2019attention}. In what follows we present a brief description of the involved attention mechanisms.

\subsection{Convolutional Multi Head Self-Attention} \label{ssec:conv_mha}

An Augmented Attention Convolutional mechanism is introduced in ~\cite{bello2019attention} which combines convolutional operations and multi-head self-attention ~\cite{vaswani2017attention}.  
Firstly, the multi-head self-attention extracts a \textit{key} (K), \textit{value} (V) and \textit{query} (Q) from each input (X), and calculates an output matrix for a head $h$ as follows:
\begin{equation}
\textrm{O}_h = \textrm{Softmax}(\frac{Q K^{T}}{\sqrt{d_k^h}})V,
\end{equation}
where 
\begin{equation}
\quad Q=XW_{q}, \quad K=XW_{k}, \quad \textrm{and} \quad V=XW_{v}.
\end{equation}
Here $d_k^h$ is the dimension of the queries per attention head. The input $X\in \mathbb{R}^{HW \times C}$ is a reshaping of the input tensor $I\in \mathbb{R}^{H \times W \times C}$. Here the height, width and the number of channels are denoted by $H$, $W$ and $C$ respectively. The outputs of all the heads are then concatenated and weighted again as follows ~\cite{vaswani2017attention}:
\begin{equation}
\textrm{MultiHead(X)} = \textrm{Concat}[\text{O}_{1}, ..., \textrm{O}_{n}]W_{o}.  
\end{equation}
Here the weights $W_q$, $W_k$, $W_v$ and $W_o$ are linear transformations that are learned during the training. Subsequently, in the Augmented Attention Convolution, a 2D convolutional operation is performed in parallel to a multi-head self-attention operation. Later, these two outputs are concatenated as follows ~\cite{bello2019attention}:
\begin{equation}
\textrm{AugAttentionConv} = \textrm{Concat}[\textrm{Conv}(X), \textrm{MultiHead}(X)].
\end{equation}
Such a mechanism produces an interaction between the input signals, allowing the model to capture longer-range dependencies. It is shown particularly useful in convolutional neural networks, which due to their nature, only extract local relations. In all the studied models in this paper, this attention mechanism is added in the first convolutional layer and the number of attention heads were set to two.

\subsection{Global Attention}

We have incorporated a Luong's style of global attention \cite{luong2015effective} into our proposed models. Given all source hidden states $\bar{h}_s$ and the current target hidden state $h_t$, one obtains the score between the two entities as follows:
\begin{equation}
score(\bar{h}_s,h_t)=h_t^TW_a\bar{h}_s,
\end{equation}
where $W_a$ are learnable weights through a feed forward network \cite{bahdanau2014neural}. For this specific use case, we consider $h_t$ as the last hidden state of the sequence while $\bar{h}_s$ are all previous hidden states. An attention score vector $a_t$ is then obtained through the dot product between $h_t$ and $score(\bar{h}_s,h_t)$, followed by a softmax function. A context vector $c_t$ is then derived as the weighted sum between $a_t$ and $\bar{h}_s$. Finally one obtains the attentional hidden state $\Tilde{h}$ as follows:
\begin{equation}
\Tilde{h}=tanh(W_c[c_t:h_t],
\end{equation}
where $W_c$ is also a learnable weight matrix. 
\section{Proposed Models}\label{sec:proposed_models}

The proposed models in this paper are based on previously studied models, i.e. EEGNet~\cite{lawhern2018eegnet}, the Cascade Network and the Multiview Network~\cite{zhang2018cascade}. These three models have been successfully applied for analyzing EEG signals and related involved classification tasks. However, their network architectures mainly rely on convolutional operations which has a significant weakness in that it only operates on a local neighborhood. Recent studies have shown that the local nature of the convolutional kernel prevents it from capturing global contexts of the given input \cite{hu2018gather}. Additionally, the two latter models, i.e. Cascade Network and Multiview Network, use LSTMs which treat all hidden states equally, while some of them might be more relevant than others. Attention mechanisms have recently emerged in the literature in order to capture long range interactions \cite{bello2019attention}. Motivated by the success of this recent advancement, this paper aims at extending the three models, EEGNet, Cascade Network and Multiview Network by incorporating several attention mechanisms and adapting their architectures to suit analyzing MEG data for cross subject multi-class classifications. It should be noted that MEG data is relatively less explored compared to EEG data. In comparison to EEG data, MEG signals are more expensive and complex to obtain, however, it is more precise due to its higher spatio-temporal resolution. In addition, in contrary to the previous models \cite{zhang2018cascade}, here the input shape of the Attention Augmented Cascade and Multiview models are also changed to tensorial format which leads to achieve a higher performance. The three proposed models work in an end-to-end fashion, from the raw input data to the final classification, therefore requiring minimum domain knowledge. In addition, the data pre-processing phase is minimal and is detailed in section \ref{sec:Preprocessing}. All the code and data used for our experiments can be found at: \href{https://github.com/SMehrkanoon/Deep-brain-state-classification-of-MEG-data}{https://github.com/SMehrkanoon/Deep-brain-state-classification-of-MEG-data}.

Here, in particular we have utilized two kinds of attention: self and global attention. Self-attention has shown to work well in convolutional layers \cite{bello2019attention}. We have augmented the first convolutional layer of the three models with self-attention mechanism. In this case, the self-attention helps to extract the most relevant spatial features. For the Cascade and Multiview networks that have recurrent layers, the global attention is used to get the most relevant temporal features of the brain signals. Since EEGNet does not have a recurrent layer, we have instead applied the global attention on the dense layer after CNN. In what follows we give the overview of the applied attention mechanisms for each model.

\subsection{Attention Augmented EEGNet (AA-EEGNet)}\label{sec:EEGNet}
The network’s architecture proposed in~\cite{lawhern2018eegnet} has served as the basis for the architecture introduced in this research. We have incorporated two types of attention mechanisms into the core architecture in order to capture the most relevant spatial and temporal features of the network. 
This model is a compact CNN, composed of three main convolutional layers. 
First, a 2D convolution, with filter size (1, K), acts as a band-pass filter. As a result, each of its filters outputs a specific range of frequency from the EEG recordings and extracts temporal features from them by convolving over the time dimension. Then, a 2D depth-wise convolution, with filter size (C, 1), convolves along the channels and thus learns spatial features. In this way, the network extracts spatial features specific for each range of frequency. 
The combination of these two convolutions acts in the same way as the Filter Bank Common Spatial Pattern (FBCSP)\cite{ang2008filter}, a popular algorithm for classifying EEG data. Lastly, a 2D separable convolution, composed of a 2D depth-wise convolution followed by a point-wise convolution, is added to the network architecture. While the first operation, with filter size (1, D), summarizes the previously extracted features, the second convolution, with filter size (1, 1), combines those features. After this sequence of convolutional operations, a softmax classifier is used to classify the extracted features. It should be noted that the first convolutional layer has been augmented with multi-head self-attention, and the global attention has been added before the softmax classifier to enhance its performance.

\begin{figure}[htbp]
\centering
\includegraphics[width=\columnwidth]{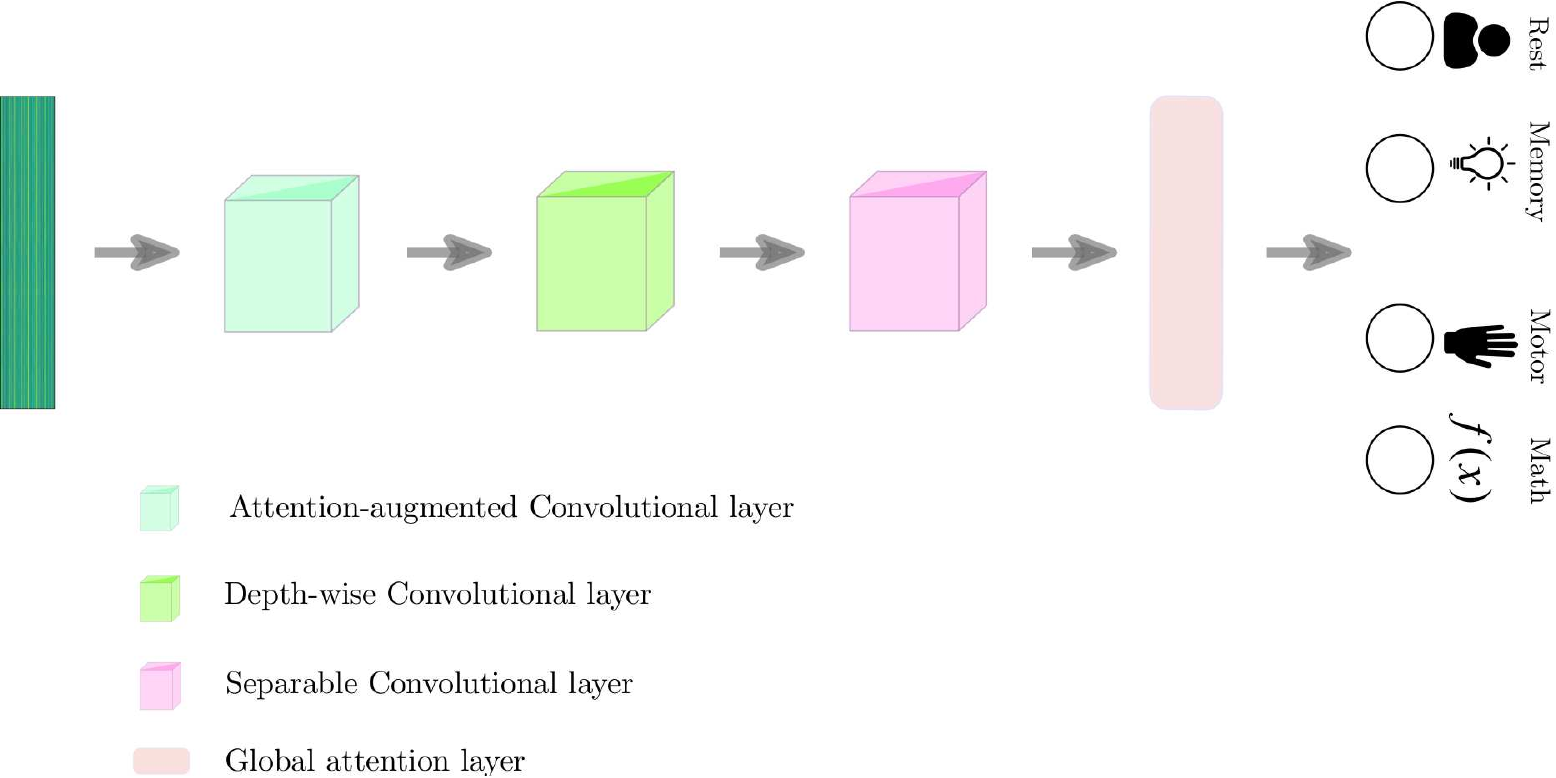}
\caption{Overall architecture of the Attention Augmented EEGNet (AA-EEGNet).}
\label{eegnet_archi}
\end{figure}

An outline of the architecture can be found in Fig. ~\ref{eegnet_archi}. 
 
Additionally, a comparison of the parameters used in ~\cite{lawhern2018eegnet} and our proposed AA-EEGNet model can be found in Table \ref{tab:hyperparamEEGNet}. Here we set to have 16 temporal (Conv), 2 spatial (DepthWiseConv), 32 point-wise filters as well as 2 attention heads with kernel length of 128. This configuration was empirically found to  perform better than other tested configurations. The number of trainable parameters of each layer used in the proposed AA-EEGNet is given in Table \ref{tab:EEGNetParams}. We can see that most trainable parameters concern the attention-augmented convolutional layer.

\begin{table}[!htbp]
\centering
\caption{Hyperparameters selection of the AA-EEGNet.}
\label{tab:hyperparamEEGNet}
\resizebox{\columnwidth}{!}{
\begin{tabular}{c c c c}\Xhline{3\arrayrulewidth}
\multirow{2}{*}{\textbf{Hyperparameter}}
&\multirow{2}{*}{\makebox[10em]{ \textbf{Vernon J. et al. \cite{lawhern2018eegnet}}}} &\multirow{2}{*}{\makebox[5em]{\textbf{Tested}}}&\multirow{2}{*}{\makebox[5em]{\textbf{Best}}}\\
 & & & \\\Xhline{3\arrayrulewidth}
Conv. filters & 8 & 8,16,32 & 16\\\hline
Depth-wise Conv. filters & 2 & 2 & 2\\\hline
Separable Conv. filters & 16 & 16,32,64 & 32\\\hline
Temporal Kernel Length  & 64 & 32,64,128,256 & 128\\\Xhline{3\arrayrulewidth}
\end{tabular}
}
\end{table}

\begin{table}[H]
\begin{center}
\caption{AA-EEGNet: layers and parameters.}
\label{tab:EEGNetParams}
\begin{tabular}{c c}\Xhline{3\arrayrulewidth}
 \multirow{2}{*}{\textbf{Layer type}} & \multirow{2}{*}{\textbf{Parameters}} \\ 
 & \\
 \Xhline{3\arrayrulewidth}
 InputLayer &  0\\
 AttentionAugmentedConv2D & 2211648\\
 BatchNormalization & 4\\
 DepthwiseConv2D & 128\\
 BatchNormalization & 8\\
 AveragePooling & 0\\
 Dropout & 0\\
 SeparableConv2D & 496 \\
 BatchNormalization & 8\\
 AveragePooling & 0\\
 Dropout & 0\\
 Flatten & 0\\
 GlobalAttentionBlock & 2112\\
 Dense & 516 \\
\Xhline{3\arrayrulewidth}
\end{tabular} \\
\end{center}
\end{table}

\subsection{Attention Augmented Cascade Network  (AA-CascadeNet)}\label{sec:Cascade}
The approach in~\cite{zhang2018cascade} is used as the backbone of our proposed model. We have incorporated the self and global attention to this model. We have also adapted the hyperparameters to suits best the analyzed MEG data. In addition, we use a different data representation for the input of this network because it empirically achieved better performance. In particular, the basis of our data representation is a mesh $M_t\in \mathbb{R}^{N \times L}$, that represents a top-down view of the human scalp at each time step $t$. This mesh is represented as follows:
\begin{gather} \label{eq:7}
M_t=\\
\nonumber
\resizebox{\columnwidth}{!}{
$
\begin{bmatrix}
\begin{smallmatrix}
0 & 0 & 0 & 0 & 0 & 0 & 0 & 0 & 0 & 0 & s_t^{121} & 0 & 0 & 0 & 0 & 0 & 0 & 0 & 0 & 0 & 0 \\
0 & 0 & 0 & 0 & 0 & 0 & 0 & 0 & s_t^{122} & s_t^{90} & s_t^{89} & s_t^{120} & s_t^{152} & 0 & 0 & 0 & 0 & 0 & 0 & 0 & 0 \\
0 & 0 & 0 & 0 & 0 & 0 & 0 & s_t^{123} & s_t^{91} & s_t^{62} & s_t^{61} & s_t^{88} & s_t^{119} & s_t^{151} & 0 & 0 & 0 & 0 & 0 & 0 & 0 \\
0 & 0 & 0 & 0 & 0 & 0 & s_t^{124} & s_t^{92} & s_t^{63} & s_t^{38} & s_t^{37} & s_t^{60} & s_t^{87} & s_t^{118} & s_t^{150} & 0 & 0 & 0 & 0 & 0 & 0 \\
0 & 0 & 0 & 0 & s_t^{177} & s_t^{153} & s_t^{93} & s_t^{64} & s_t^{39} & s_t^{20} & s_t^{19} & s_t^{36} & s_t^{59} & s_t^{86} & s_t^{117} & s_t^{149} & s_t^{176} & s_t^{195} & 0 & 0 & 0 \\
s_t^{229} & s_t^{212} & s_t^{178} & s_t^{154} & s_t^{126} & s_t^{94} & s_t^{65} & s_t^{40} & s_t^{21} & s_t^{6} & s_t^{5} & s_t^{18} & s_t^{35} & s_t^{58} & s_t^{85} & s_t^{116} & s_t^{148} & s_t^{175} & s_t^{194} & s_t^{228} & s_t^{248} \\
s_t^{230} & s_t^{213} & s_t^{179} & s_t^{155} & s_t^{127} & s_t^{95} & s_t^{66} & s_t^{41} & s_t^{22} & s_t^{7} & s_t^{4} & s_t^{17} & s_t^{34} & s_t^{57} & s_t^{84} & s_t^{115} & s_t^{147} & s_t^{174} & s_t^{193} & s_t^{227} & s_t^{247} \\
0 & s_t^{231} & s_t^{196} & s_t^{156} & s_t^{128} & s_t^{96} & s_t^{67} & s_t^{42} & s_t^{23} & s_t^{8} & s_t^{3} & s_t^{16} & s_t^{33} & s_t^{56} & s_t^{83} & s_t^{114} & s_t^{146} & s_t^{173} & s_t^{211} & s_t^{246} & 0 \\
0 & s_t^{232} & s_t^{197} & s_t^{157} & s_t^{129} & s_t^{97} & s_t^{68} & s_t^{43} & s_t^{24} & s_t^{9} & s_t^{2} & s_t^{15} & s_t^{32} & s_t^{55} & s_t^{82} & s_t^{113} & s_t^{145} & s_t^{172} & s_t^{210} & s_t^{245} & 0 \\
0 & s_t^{233} & s_t^{198} & s_t^{158} & s_t^{130} & s_t^{98} & s_t^{69} & s_t^{44} & s_t^{25} & s_t^{10} & s_t^{1} & s_t^{14} & s_t^{31} & s_t^{54} & s_t^{81} & s_t^{112} & s_t^{144} & s_t^{171} & s_t^{209} & s_t^{244} & 0 \\
0 & 0 & s_t^{214} & s_t^{180} & s_t^{131} & s_t^{99} & s_t^{70} & s_t^{45} & s_t^{26} & s_t^{11} & s_t^{12} & s_t^{13} & s_t^{30} & s_t^{53} & s_t^{80} & s_t^{111} & s_t^{143} & s_t^{192} & s_t^{226} & 0 & 0 \\
0 & 0 & 0 & 0 & s_t^{159} & s_t^{132} & s_t^{100} & s_t^{71} & s_t^{46} & s_t^{27} & s_t^{28} & s_t^{29} & s_t^{52} & s_t^{79} & s_t^{110} & s_t^{142} & s_t^{170} & 0 & 0 & 0 & 0 \\
0 & 0 & 0 & s_t^{181} & s_t^{160} & s_t^{133} & s_t^{101} & s_t^{72} & s_t^{47} & s_t^{48} & s_t^{49} & s_t^{50} & s_t^{51} & s_t^{78} & s_t^{109} & s_t^{141} & s_t^{169} & s_t^{191} & 0 & 0 & 0 \\
0 & 0 & s_t^{215} & s_t^{199} & s_t^{182} & s_t^{161} & s_t^{134} & s_t^{102} & s_t^{73} & s_t^{74} & s_t^{75} & s_t^{76} & s_t^{77} & s_t^{108} & s_t^{140} & s_t^{168} & s_t^{190} & s_t^{208} & s_t^{225} & 0 & 0 \\
0 & 0 & s_t^{234} & s_t^{216} & s_t^{200} & s_t^{183} & s_t^{162} & s_t^{135} & s_t^{103} & s_t^{104} & s_t^{105} & s_t^{106} & s_t^{107} & s_t^{139} & s_t^{167} & s_t^{189} & s_t^{207} & s_t^{224} & s_t^{243} & 0 & 0 \\
0 & 0 & 0 & 0 & s_t^{235} & s_t^{217} & s_t^{201} & s_t^{184} & s_t^{163} & s_t^{136} & s_t^{137} & s_t^{138} & s_t^{166} & s_t^{188} & s_t^{206} & s_t^{223} & s_t^{242} & 0 & 0 & 0 & 0 \\
0 & 0 & 0 & 0 & 0 & 0 & s_t^{236} & s_t^{218} & s_t^{202} & s_t^{185} & s_t^{164} & s_t^{165} & s_t^{187} & s_t^{205} & s_t^{222} & s_t^{241} & 0 & 0 & 0 & 0 & 0 \\
0 & 0 & 0 & 0 & 0 & 0 & 0 & 0 & s_t^{219} & s_t^{203} & s_t^{186} & s_t^{204} & s_t^{221} & 0 & 0 & 0 & 0 & 0 & 0 & 0 & 0 \\
0 & 0 & 0 & 0 & 0 & 0 & 0 & 0 & 0 & s_t^{237} & s_t^{220} & s_t^{240} & 0 & 0 & 0 & 0 & 0 & 0 & 0 & 0 & 0 \\
0 & 0 & 0 & 0 & 0 & 0 & 0 & 0 & 0 & 0 & s_t^{238} & s_t^{239} & 0 & 0 & 0 & 0 & 0 & 0 & 0 & 0 & 0 \\
\end{smallmatrix}
\end{bmatrix}
$
}
\end{gather}
Here, 248 MEG sensors were used, each non-zero element $s_t^j$, $j \in [1,248]$ corresponds to a specific sensor value at time step $t$. We then concatenate multiple meshes along the time axis to create a tensor $T \in \mathbb{R}^{N \times L \times D}$, where $D$ is the depth used in each of the tensors, which corresponds to the number of concatenated meshes $M_t$. As it will be shown is section \ref{ssec:results}, the depth $D$ is a hyperparameter that can in principle be tuned. Since the AA-CascadeNet is a multi-input network, we use $W$ tensors.  Thus each input of the AA-CascadeNet is a list of tensors $I_c$ $\in \mathbb{R}^{W \times N \times L \times D}$, where $W$ is the number of tensors, or input streams. $N$ and $L$ are the rows and columns of each mesh described previously, and $D$ is the depth stated above. The proposed Attention Augmented Cascade Network (AA-CascadeNet) incorporates two types of attention mechanisms which are multi-head self attention for the spatial features and global attention for the temporal features. AA-CascadeNet extracts the spatial and temporal features from the MEG signals by stacking convolution operations and recurrent units. In the first part of the model, convolutional layers take the inputs and distill the spatial features. Self-attention has been used on the first convolutional layer as it empirically showed to perform better than any other configuration. The three convolutional layers are then followed by the recurrent part of the  network, which uses two layers of LSTM units. The temporal part of the network has been augmented with global attention in order to learn to put emphasis on more important temporal features. We decided to attend the sequence between the two LSTM layers because it yielded the best accuracy compared to other tested configurations. The recurrent part of the network is then followed by a fully connected layer activated through a softmax function.  

We tested different parameters and we empirically observed that the best ones correspond to the values listed in Table \ref{tab:comparison_hyperparams_casc}, which are less than those parameters used in \cite{zhang2018cascade}.

\begin{table}[H] 
\centering
\caption{Hyperparameters selection of the Attention Augmented Cascade network.}
\label{tab:comparison_hyperparams_casc}
\resizebox{\columnwidth}{!}{
\begin{tabular}{c c c c}\Xhline{3\arrayrulewidth}
\multirow{2}{*}{\textbf{Hyperparameter}}
& \multirow{2}{*}{{ \textbf{Zhang et al. \cite{zhang2018cascade}}} } & \multirow{2}{*}{\makebox[5em]{\textbf{Tested}}}&\multirow{2}{*}{{\textbf{Best}}}\\
 & & & \\\Xhline{3\arrayrulewidth}
Filters 1st Conv. layer &32&1,2,4&1\\\hline
Filters 2nd Conv. layer &64&2,4,8&2\\\hline
Filters 3rd Conv. layer &128&4,8,16&4\\\hline
Fully connected units &1024&125,250,500&125\\\hline
Learning rate & $1e^{-4}$ & $1e^{-4},1e^{-3},1e^{-2}$ & $1e^{-4}$\\\Xhline{3\arrayrulewidth}
\end{tabular}
}
\end{table}

\begin{figure*}[]
\centering
\includegraphics[scale=0.9]{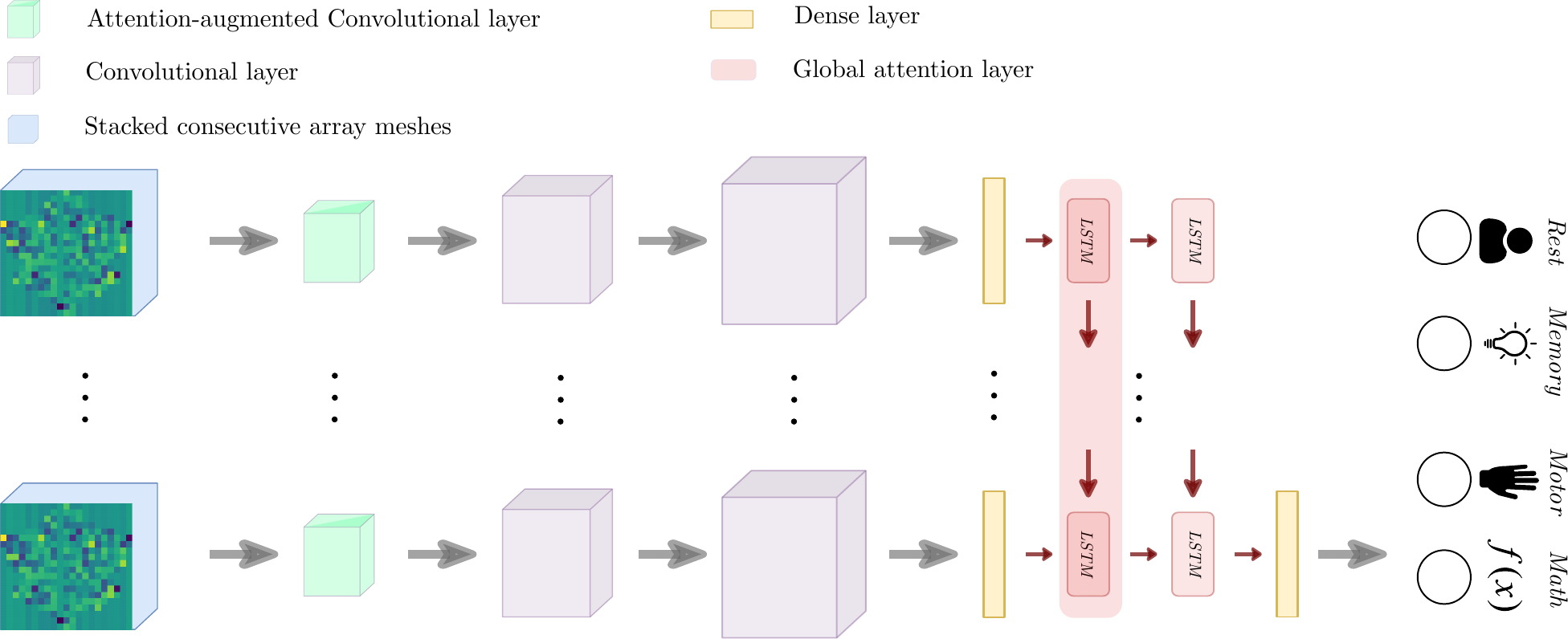}
\caption{Overall architecture of the Attention Augmented Cascade Network (AA-CascadeNet).}
\label{cascade_archi}
\end{figure*}

The maximum total number of trainable parameters in the AA-CascadeNet, which is the one that jointly uses self and global attention, is 2,139,869. Table \ref{tab:CascadeLayers} and Fig. \ref{cascade_archi} represent the structure of the model and its corresponding schema. For the sake of simplicity, in Table \ref{tab:CascadeLayers}, we only describe the network architecture for one input stream as the other streams have the same architecture. For the same reason, we don't include layers that have no learnable parameters. 

\begin{table}[!htbp]
\begin{center}
\caption{Layers and parameters of the AA-CascadeNet.}

 \begin{tabular}{c c} 
 \Xhline{3\arrayrulewidth} 
 \multirow{2}{*}{\textbf{Layer type}} & \multirow{2}{*}{\textbf{Parameters}} \\
 & \\
 \Xhline{3\arrayrulewidth} 
 SelfAttentionConv2D & 619\\
 Convolutional2D & 296\\
 Convolutional2D  & 396\\
 Dense  & 210125 \\
 concatenate & 0 \\
 LSTM  & 5440 \\
 LSTM & 840 \\
 GlobalAttention & 2660 \\
 Dense  & 16125 \\
 Dense & 504 \\
\Xhline{3\arrayrulewidth}  
\label{tab:CascadeLayers}
\end{tabular} \\

\end{center}
\end{table}

\subsection{Attention Augmented Multiview Network (AA-MultiviewNet)}
Here we propose the Attention Augmented Multiview Network (AA-MultiviewNet).
It serves the same purpose as AA-CascadeNet, which is intended to extract spatial and temporal features from brain signals in its convolutional and recurrent parts through Convolutional 2D and LSTM layers respectively. The main difference between AA-MultiviewNet and AA-CascadeNet is that in the Multiview Network, these features are obtained in parallel rather than consecutively \cite{zhang2018cascade}. The novelty introduced in this network is the same as the AA-CascadeNet, but is instead applied separately to the two streams of the network. While self-attention is used on the spatial stream of the network through an attention augmented convolutional layer \cite{bello2019attention}, global attention is applied to the temporal stream of the network. Thus, the architecture requires two types of input including a tensor list made of two-dimensional meshes of MEG channels for the spatial part and a list of matrices for the recurrent part. 

This network incorporates two streams including a spatial and a temporal. While the spatial stream uses the same list of tensors $I_c$ explained previously for the AA-CascadeNet, the input processed by the temporal part is different. We first use a vector $s_t$, that represents the sensor values of the 248 MEG channels at time step $t$. We then concatenate consecutive vectors $s_t$ to form the matrix $E \in \mathbb{R}^{G \times D}$, where $G$ is the number of MEG channels, and $D$ is the depth presented in the section above. We then use a list of matrices $I_m \in \mathbb{R}^{W \times G \times D}$ as an input for the temporal part of the AA-MultiviewNet, where $W$ is the number of tensors discussed above. It should be noted that each matrix $E$ used in the temporal part of this network has its corresponding tensor $T$ in the spatial part, meaning that the time steps used are the same and concatenated in the same order.

Starting with the spatial stream, the augmented convolutional layer is used on the first layer, as it empirically showed to lead the best result. Padding is not applied after any of the convolutional layers since the information at the edges of meshes, that can be deemed as valuable, would not be accessible otherwise. The number of kernels in the convolutional layers are doubled consecutively starting from 1 in the first layer. The shape of the kernels is (7,7) in all 3 convolutional layers.
Concerning the temporal part, each input passes through a fully-connected dense layer and the result is then concatenated to form the input sequence for the LSTM layers. The attention mechanism is applied on this input sequence through global attention. 

It should be mentioned that we also examined applying global attention in the output sequence of the first LSTM as well as the output sequence of the second LSTM, however the best results were achieved through the incorporation of global attention in the input sequence of the first LSTM layer.

Similarly to the AA-CascadeNet, here there are also two LSTM layers and the second layer does not return a sequence. After combining the outputs from the spatial and temporal parts into dense layers, they are concatenated and fed into softmax layer for final prediction.

Table \ref{tab:MultiviewParams} shows the layer configuration of the AA-MultiviewNet. CNN part starts from first layer and ends with second input layer. After that RNN part takes place until the end of second LSTM layer. Then the addition and concatenation of outputs from both parts occur. Here, for the sake of simplicity, only layers that have learnable parameters are included in Table \ref{tab:MultiviewParams}, except the addition and concatenation operations.

\begin{table}[]
\begin{center}
\caption{Layers and parameters of the AA-MultiviewNet.}
\label{tab:MultiviewParams}
 \begin{tabular}{c c} 
 \Xhline{3\arrayrulewidth}
 \multirow{2}{*}{\textbf{Layer type}} & \multirow{2}{*}{\textbf{Parameters}} \\ 
 & \\ 
 \Xhline{3\arrayrulewidth}
 SelfAttentionConv2D & 619\\
 Convolutional2D & 296\\
 Convolutional2D  & 396\\
 Dense  & 210125  \\
 Add & 0 \\
 Dense & 31125 \\
 LSTM  & 5440 \\
 GlobalAttention & 2660 \\
 LSTM & 480 \\
 Dense & 1375 \\
 concatenate & 0 \\
 Dense & 1004 \\
\Xhline{3\arrayrulewidth} 
\end{tabular} \\
\end{center}
\end{table}

\begin{figure}[htbp]
\centering
\includegraphics[width=\columnwidth]{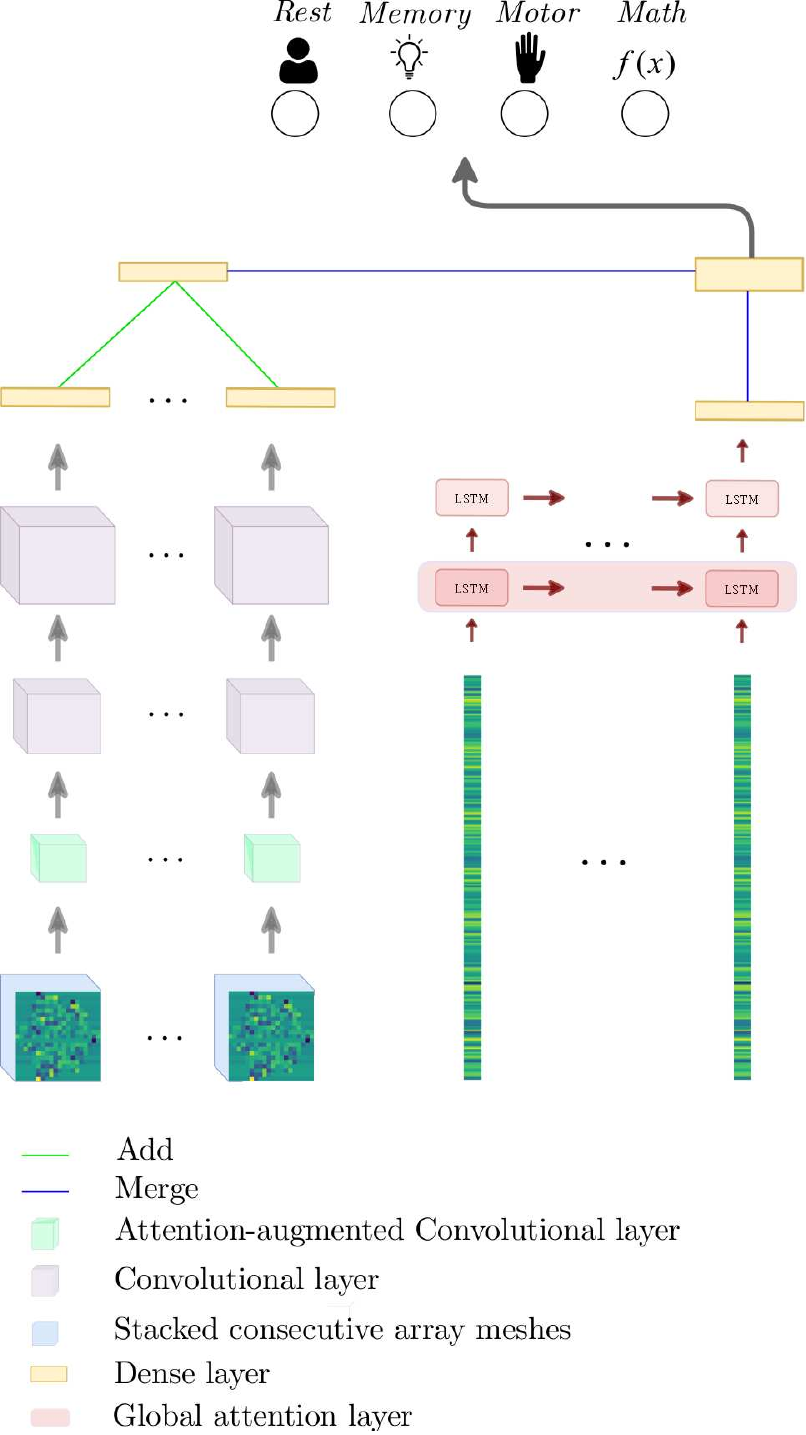}
\caption{Overall architecture of the Attention Augmented Multiview Network (AA-MultiviewNet).}
\label{multiview_archi}
\end{figure}

\section{Data Description}\label{sec:data_desc}

MEG data was collected from the publicly available database of the Human Connectome Project (HCP)\cite{van2012human}. To record the data, all subjects from the HCP underwent scanning using a whole head MAGNES 3600 by 4D Neuroimaging. This system includes 248 magnetometer channels and 23 reference channels. A sampling rate of 2034.5101 Hz was used to record the data. The full specifications of the system can be found in the official documentation of the HCP\footnote{\url{https://www.humanconnectome.org/study/hcp-young-adult/document/1200-subjects-data-release}}.

In this paper, the dataset called \textit{1200 Subjects Release (S1200)} and released by the WU-Minn HCP consortium was used. Out of the 1200 young adult (ages 22-35) subjects in the dataset, 95 have MEG data. After further filtering out subjects, whose data has some issues such as being severely lacking in quantity compared to the rest, or having channels that do not operate properly, we ended up working with eighteen subjects. We use twelve of them as training and validation data, and six as test data. The subjects were in 4 different states during the recording, therefore requiring activity from different cortical and sub-cortical brain regions. In what follows we detail the different classes.

\begin{itemize}
\item \textbf{Resting state} : During 3 runs of 6 minutes each, subjects are asked to relax in a supine position with their eyes open, and to maintain visual contact on a red crosshair with dark background.
\item \textbf{Story vs. Math state} : These sessions consist of 2 runs of 7 minutes each and were divided into blocks. During the blocks that concern language processing, subjects are presented with audio short stories extracted from Aesop's fables and are then asked to answer questions about the comprehension of the short stories. During the math blocks, subjects perform arithmetic operations (addition and subtraction).
\item \textbf{Working Memory state} : These sessions last 2 runs of 10 minutes each. Each run is divided into multiple blocks and half of the blocks uses a 2-back working memory task (comparing the current picture with the one that was seen 2 pictures before) while the other half uses a 0-back memory task, used as a working memory comparison. The pictures consisted of faces or tools, and the subjects were asked to evaluate each picture, pressing one of 2 buttons depending whether the images matched or not.
\item \textbf{Motor state} : During 2 runs of 14 minutes each, subjects are presented with visual cues they need to reproduce using either the right hand, left hand, right foot, or left foot.
\end{itemize}
For more details on the specific protocol used, please refer to the official documentation\footnote{
\url{https://www.humanconnectome.org/storage/app/media/documentation/s1200/HCP_S1200_Release_Reference_Manual.pdf}}.

\section{Experimental Results}\label{sec:results}

\subsection{Data Preprocessing}\label{sec:Preprocessing}
\begin{figure}[!htbp]
\centering
\includegraphics[width=\columnwidth]{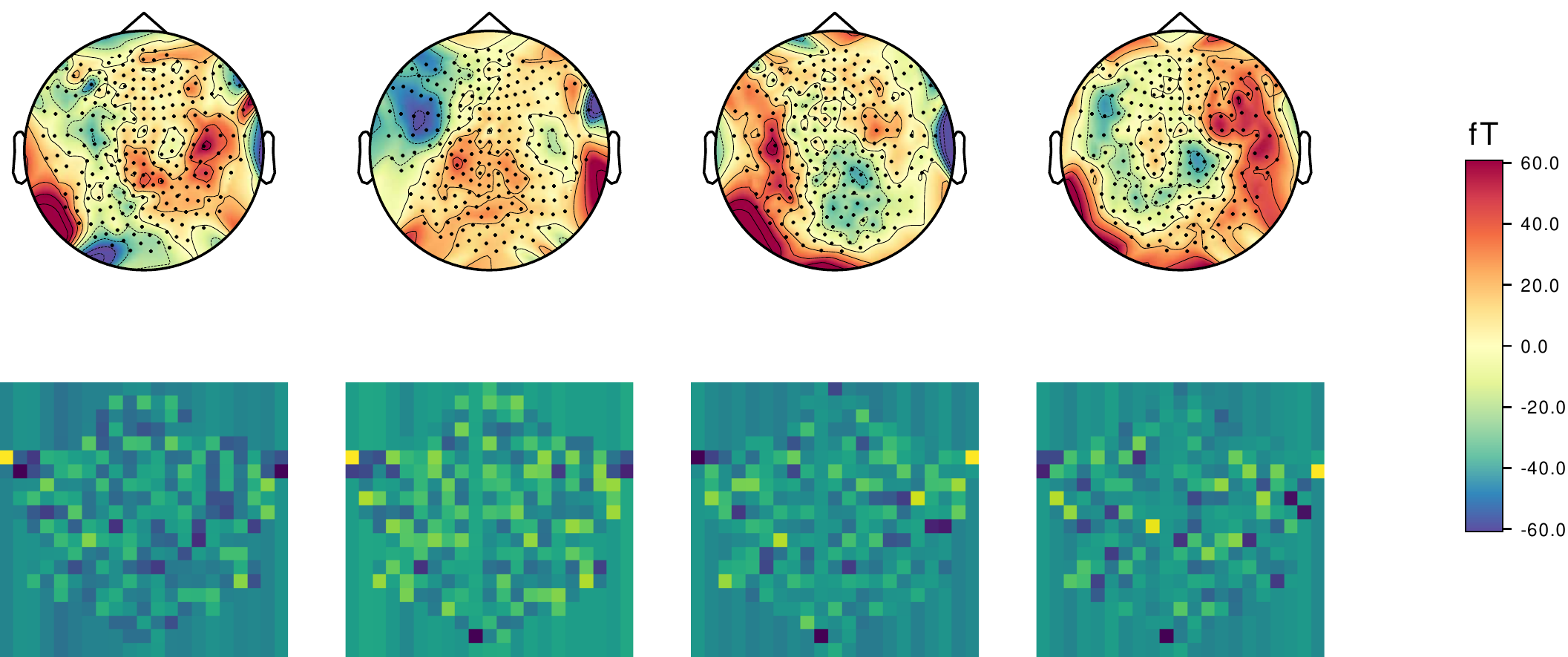}
\caption{Example of raw data captured from the magnetometers and their corresponding mesh representation after preprocessing.}
 \label{fig:input_topo}
\end{figure}

A minimal preprocessing step is done on the raw data. As stated previously, the order of magnitude of MEG data is very small and needs preprocessing to be fed to the neural networks. For the Attention Augmented Cascade and Multiview networks, a data normalization was performed. In the AA-EEGNet, a scaling (i.e. $ \times10^5 $ factor) is performed. This scaling factor was empirically found to lead the best performance compared to other scaling factor candidates for the studied dataset. The spatial data representation used for the multi-input networks are meshes  that represent a top-down view of the human scalp according to where the channels are connected to the subject's brain. An example of such inputs can be found in Fig.  \ref{fig:input_topo}. This process was done under the same assumption as in \cite{zhang2018cascade}, where specific tasks activate distinct areas of the brain and these meshes allowed to include spatial information about the 248 MEG channels as an input to the convolutional layers. In this mesh representation, all values outside of the channels were set to 0. This mesh is shown in \ref{eq:7}.

For the AA-EEGNet, the input of the network was made of segments of 1425 time steps each. This window size is the default one in the MNE toolbox \cite{gramfort2013meg} used to split the recording.

\begin{figure}[H]
\centering
\includegraphics[width=2.5in]{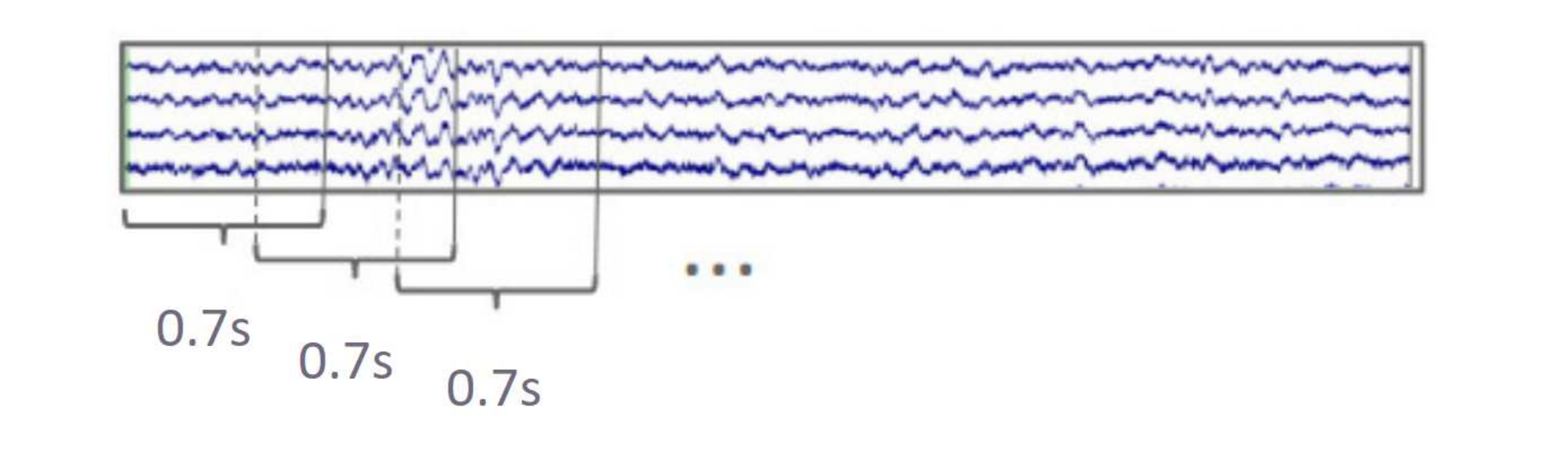}
\caption{Segmentation of the brain signals for the AA-EEGNet.}
\label{fig:segmentation_eegnet}
\end{figure}

Regarding the segmentation of the data, we used different lengths depending on the networks. With the AA-EEGNet, a  sliding window of 0.7s with a 33\% of overlapping between each window over the MEG recordings empirically found to yield the best results. In Fig.  \ref{fig:segmentation_eegnet}, this segmentation can be seen graphically.
Concerning the AA-CascadeNet and AA-MultiviewNet, a 50\% overlapping between the number of input streams, i.e. $W$, across all data samples is used. More precisely, if we consider two data samples and an even number of input streams $W$, then the second half of the input streams of the first data sample is the same as first half of the input streams of the second data sample. Fig. \ref{fig:input_topo} shows one of the meshes used for each class, given as input to the AA-CascadeNet and the spatial input part of the AA-MultiviewNet. From left to right, we can look at data extracted during rest, memory, math, and motor tasks.
 
\subsection{Experimental setup}
The experimental setup has been the same for all the networks. The experiments are divided into two setups where each contained a different number of training subjects. The first and second setup consists of 3 and 12 training subjects, respectively. The test subjects are kept the same in both setups. Such scenarios were selected in order to asses how well the models can generalize across different groups of subjects as well as how adding more training subjects can influence the performances of the models. Since we wanted to investigate how well the models can generalize over multiple unseen subjects, six subjects are used for the test phase.

\subsection{Training} \label{ssec:training}
Adam method \cite{kingma2014adam} is used to optimize the categorical cross-entropy with a learning rate of $1e^{-4}$ and a batch size of 64 in the case of the Attention Augmented Cascade and Multiview networks. The AA-EEGNet uses the same parameters except for the batch size, which is 16. The activation function used in the AA-Cascade and AA-Multiview networks is the rectified linear unit (ReLu) while the exponential linear unit (Elu) function was used for AA-EEGNet. 
\subsection{Results} \label{ssec:results}
The obtained mean accuracy and the standard deviation of the various models over the six test subjects are tabulated in Table \ref{tab:results_all_models}. For each applied architecture, we show the setup it was trained on, as well as the attention configuration.

\begin{table}[H]
    \centering
    \renewcommand{\arraystretch}{1.5} 
    \caption{Cross subject classification results of the 3 models.}
    \label{tab:results_all_models}
    \resizebox{\columnwidth}{!}{
    \begin{tabular}{c c c c c}\Xhline{3\arrayrulewidth}
    \multirow{2}{*}{\textbf{Model}}&\multirow{2}{*}{\textbf{Setup}} & \multirow{2}{*}{\makebox[7em]{\textbf{No Attention}}} & \multirow{2}{*}{\makebox[7em]{\textbf{Self Attention}}} & \multirow{2}{*}{\makebox[10em]{\textbf{Self + Global Attention}}}\\
     & & & & \\\Xhline{3\arrayrulewidth}
         \multirow{2}{*}{AA-EEGNet} & 1 & 0.74 $\pm$ 0.17 & 0.72 $\pm$ 0.20 & 0.80 $\pm$ 0.06 \\
         & 2 & 0.83 $\pm$ 0.15 & 0.90 $\pm$ 0.08 & 0.90 $\pm$ 0.08 \\ \hline
         \multirow{2}{*}{AA-CascadeNet} & 1 & 0.76 $\pm$ 0.11 & 0.88 $\pm$ 0.08 & 0.70 $\pm$ 0.18\\
         & 2 & \underline{0.91 $\pm$ 0.07} & 0.92 $\pm$ 0.08 & \underline{0.93 $\pm$ 0.06}\\ \hline 
         \multirow{2}{*}{AA-MultiviewNet} & 1 & 0.71 $\pm$ 0.14 & 0.68$\pm$ 0.07 & 0.71 $\pm$ 0.08\\
         & 2 & 0.90 $\pm$ 0.07 & \underline{0.94 $\pm$ 0.07} & 0.92 $\pm$ 0.08\\\hline\hline
        \multicolumn{1}{c}{\multirow{2}{*}{The highest Accuracy}}&& \multirow{2}{*}{0.91 $\pm$ 0.07} &\multirow{2}{*}{0.94 $\pm$ 0.07} & \multirow{2}{*}{0.93 $\pm$ 0.06}\\
         &  &  & \\
        \Xhline{3\arrayrulewidth}
    \end{tabular}
    }
\end{table}

From Table \ref{tab:results_all_models}, in general one can observe the benefit of incorporating the attention mechanism in improving the generalization performance of the models across multiple subjects. For some models this improvement is more significant than other. In particular, the positive influence of the attention mechanism on the EEGNet can clearly be seen for setup 2. From Table \ref{tab:results_all_models}, it can also be seen that as the number of training subjects increases, i.e. going from setup 1 to setup 2, the accuracy is improved and in general the influence of the added attention mechanisms is more pronounced. 
Overall, the AA-MultiviewNet with self-attention yields the best result, with the maximum accuracy of $94\%$. In addition, when considering the performances related to setup 1, using the attention mechanism does not consistently give better results than using the vanilla models.
One can explain these results under two assumptions. The first one is the lack of samples in setup 1, preventing the attention mechanism to learn important features, and thus to be effective as generalizing to unseen subjects. The second one is the increased complexity of the attention models. Since the attention-augmented convolutional layer requires additional intermediate convolutions as well as non-trivial operations like tiling, transposing, and reshaping \cite{bello2019attention}, we think it negatively impacts the performance when given a smaller number of samples.

The results corresponding to different number of depths $D$, used to construct the explained input tensors in section \ref{sec:Cascade}, for AA-CascadeNet and AA-MultiviewNet are tabulated in Tables \ref{tab:results_channels_cascade} and \ref{tab:results_channels_multiview} respectively. 

\begin{table}[H]
    \centering
    \renewcommand{\arraystretch}{1.5} 
    \caption{Cross subject classification results when using 3 different depths for the AA-CascadeNet and for setup 2.}
    \label{tab:results_channels_cascade}
    \resizebox{\columnwidth}{!}{
    \begin{tabular}{c c c c}\Xhline{3\arrayrulewidth}
    \multirow{2}{*}{\textbf{Depth}}& \multirow{2}{*}{\makebox[7em]{\textbf{No Attention}}} & \multirow{2}{*}{\makebox[7em]{\textbf{Self Attention}}} & \multirow{2}{*}{{\textbf{Self + Global}}}\\
     & & & \\\Xhline{3\arrayrulewidth}
         1 & 0.88 $\pm$ 0.07 & 0.78 $\pm$ 0.11 & 0.74 $\pm$ 0.12\\
         10 & 0.88 $\pm$ 0.08 & 0.91 $\pm$ 0.07 & \underline{0.93 $\pm$ 0.06}\\
         100& \underline{0.91 $ \pm$ 0.07} & \underline{0.92 $ \pm$ 0.08} & 0.93 $ \pm$ 0.08 \\
        \Xhline{3\arrayrulewidth}
    \end{tabular}
    }
\end{table}

Table \ref{tab:results_channels_cascade} shows the influence of the depth used in the input tensors of the AA-CascadeNet. One can notice that when using a depth of 1, the attention mechanism is ineffective.  

\begin{table}[!htbp]
    \centering
    \renewcommand{\arraystretch}{1.5} 
    \caption{Cross subject classification results when using 3 different depths for the AA-MultiviewNet and for setup 2.}
    \label{tab:results_channels_multiview}
    \resizebox{\columnwidth}{!}{
    \begin{tabular}{c c c c}\Xhline{3\arrayrulewidth}
    
    \multirow{2}{*}{\textbf{Depth}}& \multirow{2}{*}{\makebox[7em]{\textbf{No Attention}}} & \multirow{2}{*}{\makebox[7em]{\textbf{Self Attention}}} & \multirow{2}{*}{{\textbf{Self + Global}}}\\
     & & & \\\Xhline{3\arrayrulewidth}
         1 & 0.88 $\pm$ 0.07 & 0.84 $\pm$ 0.09 & 0.87 $\pm$ 0.08\\
         10& \underline{0.90 $\pm$ 0.07} & \underline{0.94 $\pm$ 0.07} & 0.90 $\pm$ 0.07\\
         100& \underline{0.90 $ \pm$ 0.07} & 0.92 $ \pm$ 0.07 & \underline{0.92 $ \pm$ 0.08} \\
        \Xhline{3\arrayrulewidth}
    \end{tabular}
    }
\end{table}
Table \ref{tab:results_channels_multiview} shows that using the depths of ten and a hundred is more beneficial to the network than using a depth of one. In terms of consistency, using a depth of ten  is better as it yields a constant standard deviation of 7\%. In addition, self-attention greatly benefits from the depth of ten, with an accuracy of 94\%.
\begin{figure}[H]
\centering
\includegraphics[width=\columnwidth]{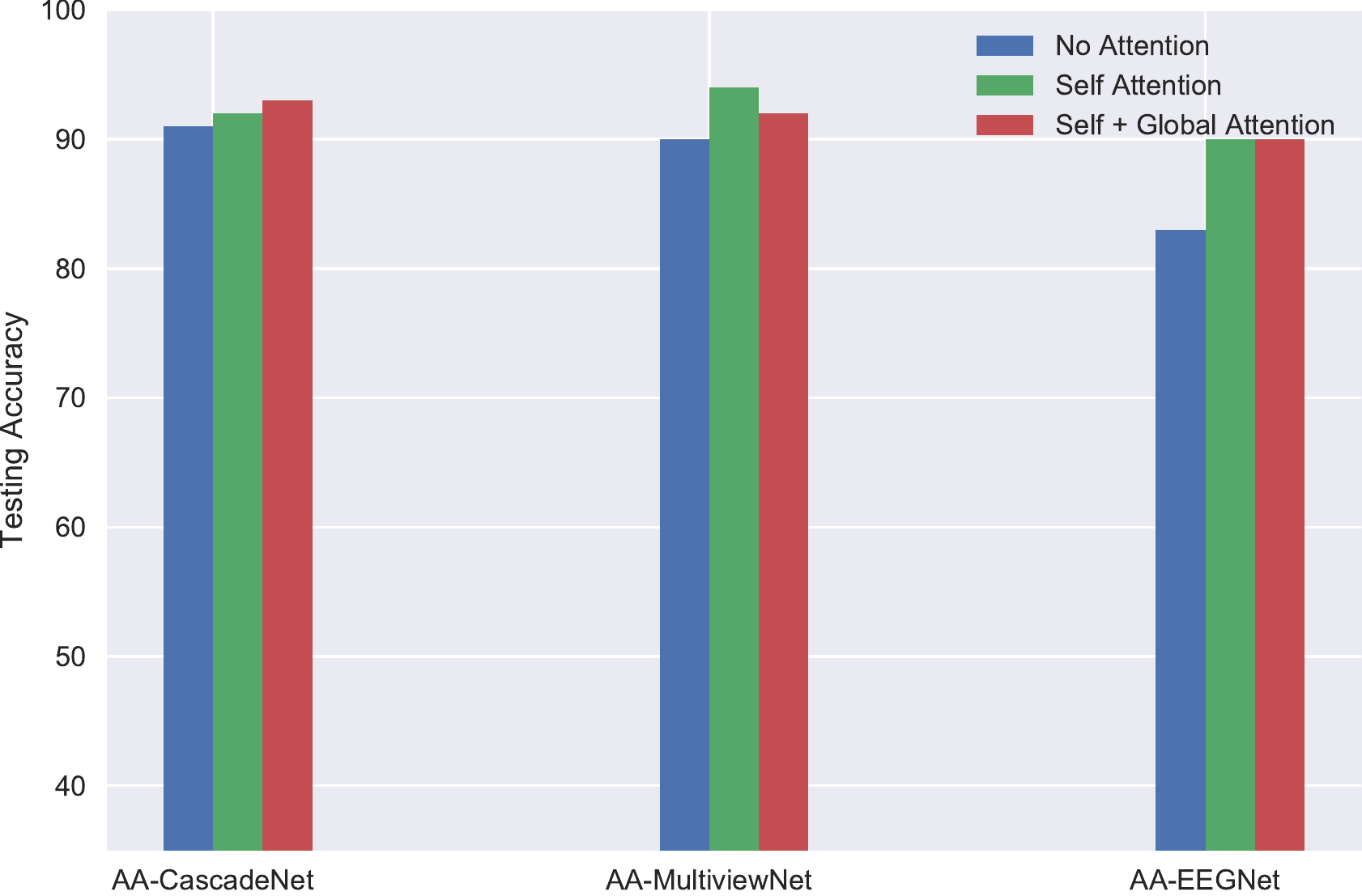}
\caption{\label{fig:comp_models_hist}Model comparison in terms of test  accuracy.}
\end{figure}

A performance comparison of the models with the relation to the use of attention mechanisms for the setup 2 are shown Fig. \ref{fig:comp_models_hist}.
As it can been seen from Fig. \ref{fig:comp_models_hist}, there is a clear benefit of incorporating the attention mechanism into the models. Depending on the used core model, self or a combination of self and global attention yields the best results.   
\begin{figure}[H]
\centering
\includegraphics[width=\columnwidth]{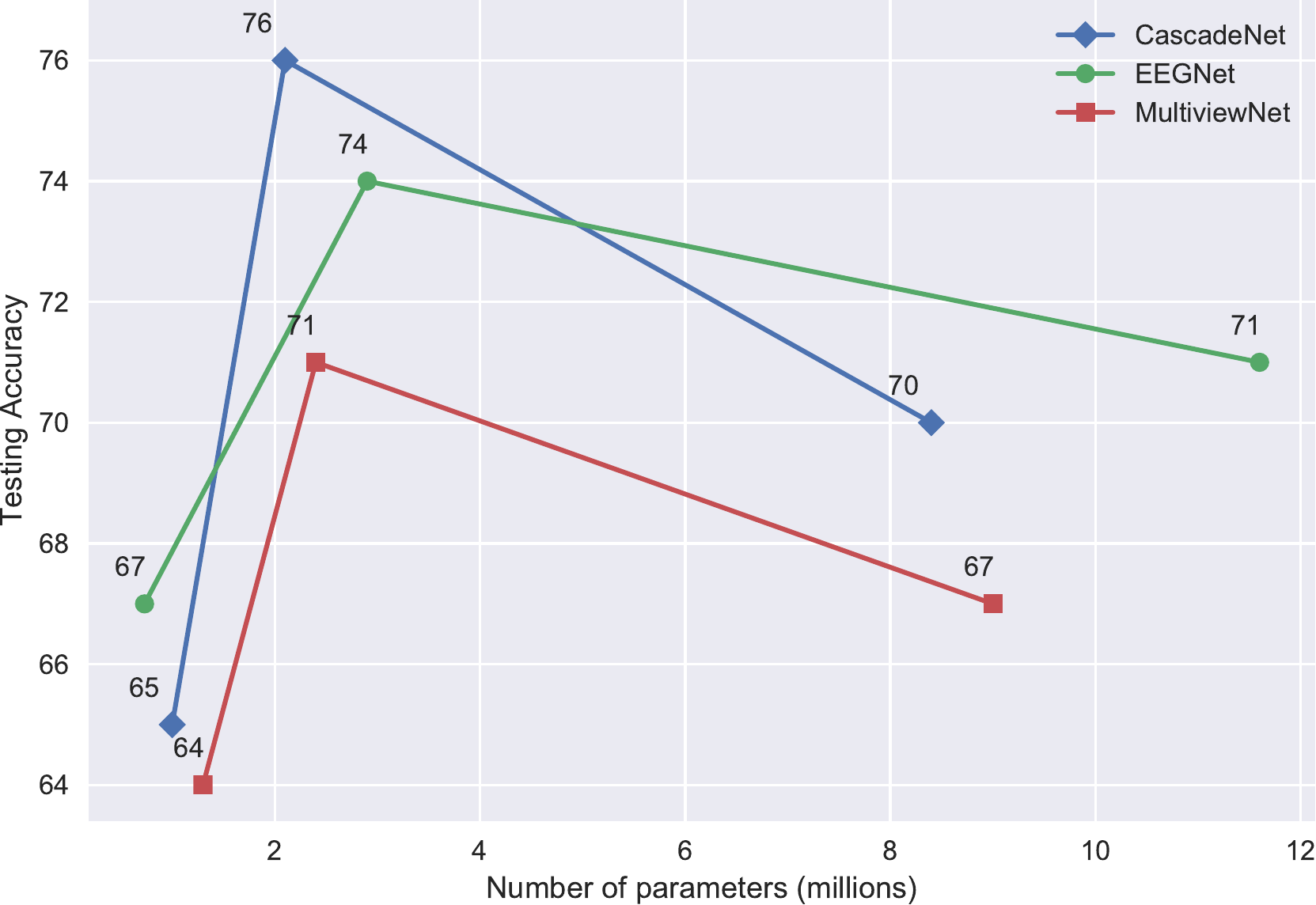}
\caption{\label{fig:accuracy_vs_params}Test accuracy in function of the number of parameters for EEGNet, the CascadeNet and the MultiviewNet.}
\end{figure}

In Fig. \ref{fig:accuracy_vs_params}, we also analyze how the total number of trainable parameters affect the performance of the models on the test set. These experiments were made on the core models, without attention. For the CascadeNet and the MultiviewNet, the number of filters  and fully connected nodes were doubled. The same process applies to EEGNet, except the number of depth-wise filters.
\section{Discussion}\label{sec:discussion}

\subsection{AA-EEGNet}

\begin{figure}[H]
    \centering
    \includegraphics[width=\columnwidth, trim={0 10.5cm 0 10.5cm},clip]{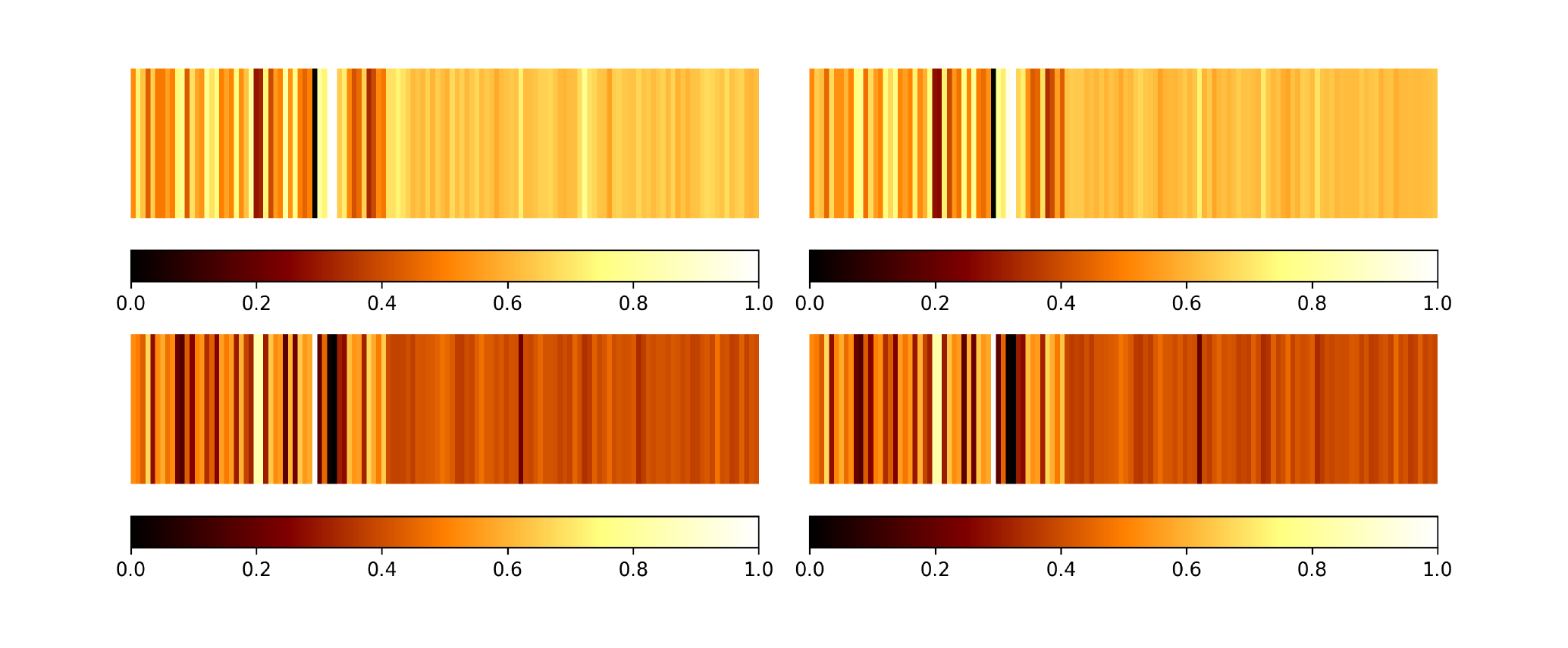}
    \caption{Augmented Convolutional filters of the Attention Augmented EEGNet  for a depth of 1.}
    \label{fig:attentionMapsEEGNet}
\end{figure}

From the results, we can observe that the addition of multi-head self-attention mechanism is making the network learn more relevant features from the input signal and thus perform more accurate predictions. The self-attention  mechanism is placed in the first convolutional layer, where the signal is separated into different ranges of frequency, and the extraction of temporal features is carried out. As seen in section \ref{ssec:conv_mha}, this kind of attention allows the various inputs to interact with each other, giving more weight to the most relevant parts of the input signal. A demonstration of such a mechanism can be found in Fig \ref{fig:attentionMapsEEGNet}, were some filters of the augmented convolution are shown. The two visualizations at the top of the figure correspond to the first filters of the self-attention operation. In these two filters, we can notice that a larger portion of them has brighter colours (higher value) than the two filters below. These two filters correspond to some of the last filters of the layer, and don't incorporate self-attention mechanism. That means that the first filters are giving more importance to some parts of the input signal than the last ones, or, in other words, the ranges of frequencies extracted by the first filters are receiving more importance than the ranges extracted by the last filters. That is a helpful mechanism, that doesn't call for domain expertise, to identify which frequencies contain more meaningful information.

Also, from table \ref{tab:results_all_models}, we can appreciate an improvement in the performance of the models with attention, especially when more subjects are added to the training dataset.

\subsection{AA-CascadeNet}
\begin{figure}[H]
\centering
\includegraphics[width=\columnwidth]{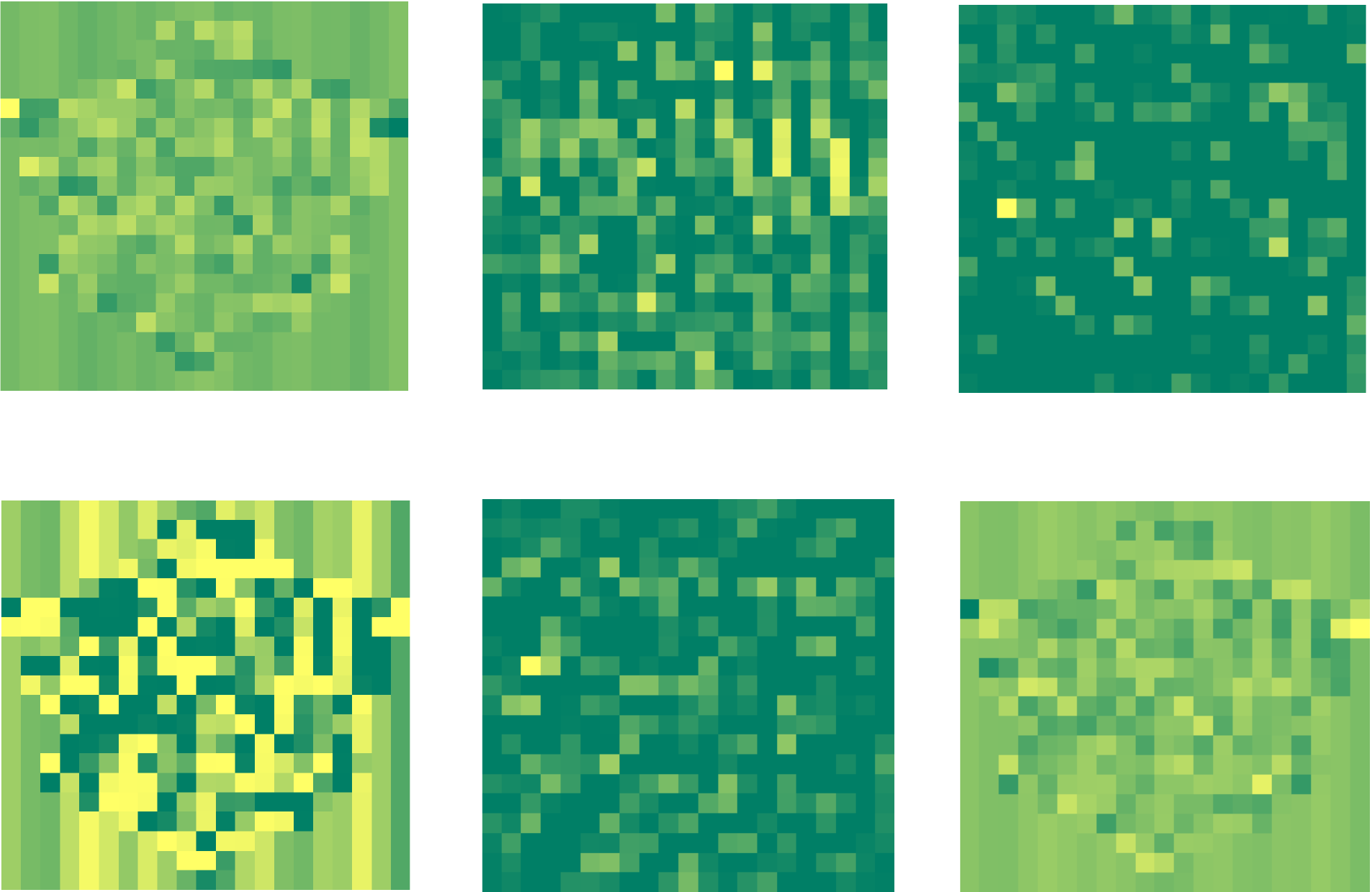}
\caption{Feature maps of the AA-CascadeNet and CascadeNet trained on setup 2. The top and bottom row shows the results without attention and with attention, respectively. From left to right, the feature maps belong to the 1st, 2nd and 3rd layers, respectively.}
\label{featuremaps_casc}
\end{figure}

As can be seen in table \ref{tab:results_all_models}, attention mechanism has a clear benefit on the model. There is a great progression in terms of accuracy, i.e. $91\%$, $92\%$, and $93\%$ for the model without attention, with self-attention only and the one using a combination of self and global attention, respectively. Additionally, the latter model achieves higher consistency with the minimum standard deviation of $6\%$ compared to all other standard deviations across all models and all setups.
Plotting the feature maps of the networks can help us better understand how the networks process the inputs. Fig. \ref{featuremaps_casc} shows a comparison of the feature maps with and without attention. Those are feature maps after every single convolutional layer for the $5^{th}$ input tensor, for a $motor$ data type.

First, it can be seen that the model without attention makes the feature maps sparser as we go deeper into the network, which is the opposite for the model with attention. This model is capable of preserving the original diamond-like shape throughout the network even if the shape seems lost in the second layer. Another expected, yet interesting finding is the high contrast of the feature maps on the first layer of the model with attention, showing that two heads are enough to extract meaningful spatial features.

\begin{figure}[H]
\centering
\includegraphics[width=\columnwidth]{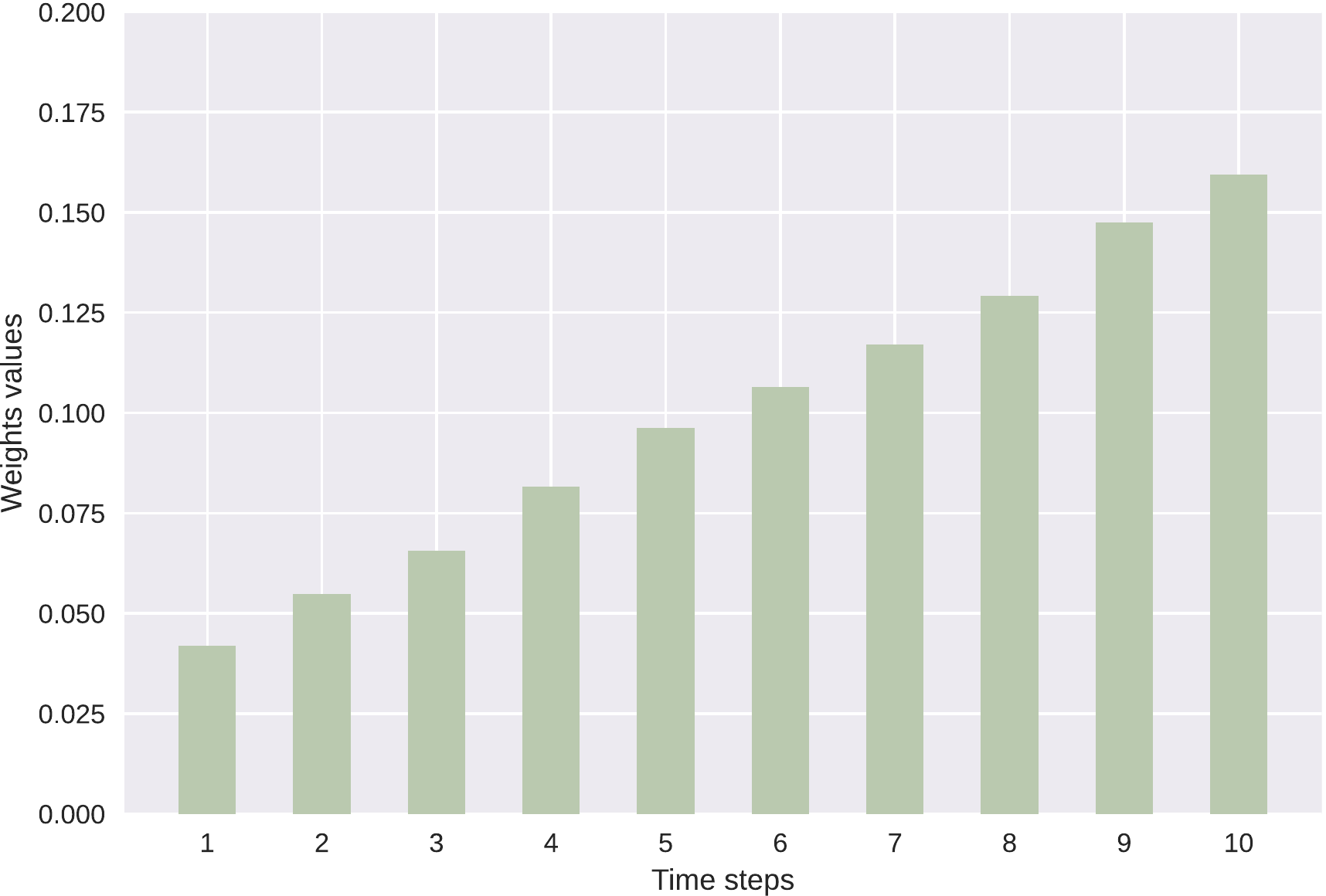}
\caption{Average global attention weights in the AA-CascadeNet trained on setup 2. }
\label{fig:weights_cascade}
\end{figure}

The global attention weights used in the Attention Augmented Cascade network are shown in Fig. \ref{fig:weights_cascade}. Since the global attention has been incorporated in the output sequence of the first LSTM layer and that the hidden size was set to 10, the sequence length is of size 10. This plotting helps us understand which part of the network seems more important at making the prediction. The last time steps seem more relevant than the first ones since their weight values are higher.

\subsection{AA-MultiviewNet}
Based on the overall results, it can be concluded that increasing the number of training subjects has a high impact on model's performance just like the other architectures. This model yielded the best performance $(94\%)$ compared to the other models, across all training setups and network configurations.
From Table \ref{tab:results_all_models}, it can clearly be observed that the model with only self-attention performs the best in terms of accuracy. We can also notice that the same type of attention is detrimental to the network when trained of setup 1, giving an accuracy of 68\%, compared to the accuracy of 71\% for both of the other network configurations. As explained previously, we assume this is due to the lack of training data for setup 1. 
\begin{figure}[H]
\centering
\includegraphics[width=\columnwidth]{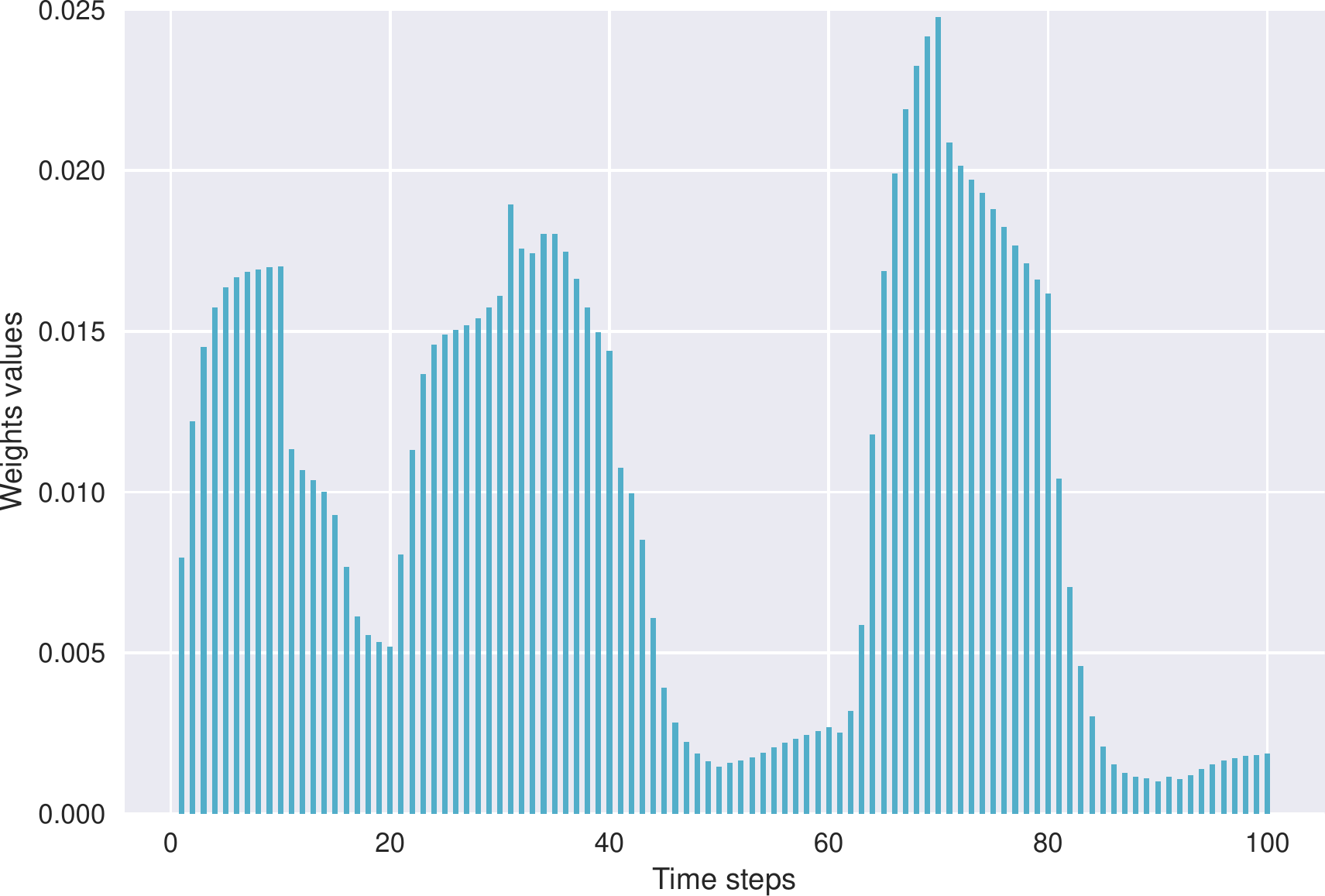}
\caption{Average global attention weights from 500 data samples in the AA-MultiviewNet trained on setup 2 and using a depth of 10.}
\label{fig:weights_mv}
\end{figure}
Fig. \ref{fig:weights_mv} shows the global attention weights used in the AA-MultiviewNet with a depth of 10. In the temporal stream of the network, ten inputs where used and after concatenating each of these inputs, we end up with a sequence length of 100. The network learned to put more emphasis on three areas, and they correspond to inputs number 1,3,4,7, and 8.

\section{Conclusion}\label{sec:conclusion}
In this paper three new architectures have been proposed and examined for the cross subject, multi-class classification tasks using MEG data. The core architectures are based on previous studied models in the literature applied to EEG data. The proposed models are equipped with two types of attention mechanisms which showed to have a clear benefit on improving the generalization performance of the models on multiple test subjects. More specifically, jointly using global attention and multihead self-attention on specific areas of the network can lead to greater performance. Moreover, our end-to-end methodology can achieve good accuracy without the need of domain knowledge, making MEG decoding a more approachable field for data practitioners. Finally, this paper also shows a potential good model transferability between EEG decoding and MEG decoding, even though the nature of the data is different. The data and code used can be found at \url{https://github.com/SMehrkanoon/Deep-brain-state-classification-of-MEG-data}.

\section*{Acknowledgment}
The present paper used RWTH Aachen's HPC cluster infrastructure for the majority of the experiments.


\bibliography{main}

\end{document}